# LEARNING TO PRONOUNCE AS MEASURING CROSS-LINGUAL JOINT ORTHOGRAPHY-PHONOLOGY COMPLEXITY

Domenic Rosati[1]

[1]scite.ai Brooklyn, New York, USA
dom@scite.ai

***ABSTRACT***

*Machine learning models allow us to compare languages by showing how hard a task in each language might be to learn and perform well on. Following this line of investigation, we explore what makes a language "hard to pronounce" by modelling the task of grapheme-to-phoneme (g2p) transliteration. By training a character-level transformer model on this task across 22 languages and measuring the model's proficiency against its grapheme and phoneme inventories, we show that certain characteristics emerge that separate easier and harder languages with respect to learning to pronounce. Namely the complexity of a language's pronunciation from its orthography is due to the expressive or simplicity of its grapheme-to-phoneme mapping. Further discussion illustrates how future studies should consider relative data sparsity per language to design fairer cross-lingual comparison tasks.*

***KEYWORDS***

*Phonology, Orthography, Linguistic Complexity, Grapheme-to-Phoneme, Transliteration*

## 1. INTRODUCTION

The aim of this work is an initial computational exploration of what makes a language "hard to pronounce" as suggested in the discussion for the SIGMORPHON 2020 shared task of grapheme-to-phoneme (g2p) transliteration [1]. Specifically, we will look at learning to transliterate several languages into the international phonetic alphabet (IPA), a form of g2p transliteration, as an indicator of language complexity. In order to do this, we train a character-level transformer model on g2p transliteration for 22 languages and measure the ratio of grapheme-to-phoneme used in languages against the character-level accuracies per word that is achieved by training the models under medium-resource assumptions. In the analysis below, we show that this method can empirically demonstrate intuitive notions of cross-lingual complexity such as phonetic orthographies being easier to learn than non-phonetic ones. We also show that data sparsity is a key issue in making cross-lingual comparisons that hasn't been well addressed in previous literature. Specifically, we will argue that it is unfair to compare languages with different inventories using the same training and testing set sizes.

Transliteration is defined here as the process of finding a mapping from one source language orthography to a target orthography. Sometimes the target transliteration alphabet is a simplification or variation of the source language orthography such as using a simplified alphabet, or an orthography designed to elucidate pronunciation as can be found in simplified English pronunciation schemes. Other times transliteration is used to map language artifacts native to one language to a transliteration that elucidates its pronunciation in another alphabet for readers of the target language. For example, in romanization a transliteration of a name natively belonging to one language is written in a "roman" or Latin-based script such as English in order to be readable by a reader of a Latin script or is written under constraints such as only having access to a Latin script-based keyboard. In this work we specifically focus on the g2p transliteration task that preserves a source language's native orthography and outputs letters drawn from IPA. It is our belief that setting up the task in this way ensures that we are learning informative end-to-end pronunciation transliterations that are useful for comparing languages. Previous g2p task set ups have allowed human or automated romanization of source orthographies (see descriptions of this in [1]) which we believe is detrimental to measuring language complexity. This is particularly important in orthographies such as used by Mandarin where romanization such as in Pinyin is giving the language a "phonetic" head start. Since we don't allow English to be reduced to a simpler phonetic variety, why should we allow other languages to be?

## 2. Previous Work

For natural language processing, transliteration models provide many unique and desirable qualities for inspecting language learning. First g2p language models have had a particular historical importance due to their role in speech to text and automatic speech recognition systems. Second, phonology and orthography are well studied areas. Theoretical models about how sequences of characters should be composed orthographically and how pronunciation occurs are readily available for researchers to guide their modelling of the task of transliteration including how feature representations, loss, and evaluation functions should be designed in order to preserve features such as phonetic similarity [2]. Third, unlike the high dimensionality of word and audio feature spaces, transliteration models have relatively low complexity in their feature spaces as they are bounded by comparatively low alphabet sizes. Because of this transliteration models may provide interesting opportunities for inspecting tasks that traditionally would have required more costly audio inputs or outputs. Additionally, the transliteration process is itself interesting as humans commonly need to engage in understanding and explaining the pronunciation of written texts and the process of learning to transliterate requires some competency with the source and target orthographies that may elucidate the dynamics of the language that is being modelled. Finally, unlike other language modelling tasks, transliteration models often focus on achieving good results in low resource scenarios where 1k (low resource) or 10k (medium resource) training samples are available. Insights for modelling with little data then may be drawn from work in transliteration.

In our work, one aspect that we'd like to study is what transliteration models can tell us about language complexity. Language complexity is a well-studied field that asks whether some languages or dimensions of language such as its syntax or phonology can be more complex or harder to learn than others [3]. Typically, this is based on classical linguistic analysis of inventories of the number of features and rules available within a phonology, orthography, or syntax system. Recent studies, notably [4] and [5], have proposed that cross-lingual complexity can be framed as a learning problem: How hard is it to learn a model of reading, writing, or pronouncing? Based on the accuracy of a particular modelling applied across languages we can potentially draw empirical conclusions of how complex one language might be with respect to another. In this respect, [4] is the closest analysis to this work that tries to understand how hard one language might be to transliterate (grapheme-to-phoneme) or decode (phoneme-to-grapheme). In our work, we look particularly at g2p transliteration to draw conclusions about joint orthography and phonology complexity.

## 3. Methods

For our g2p transliteration task, we chose to use the ipa-dict dataset of which consists of monolingual wordlists with IPA transliteration in 28 languages [6]. The alternative dataset under consideration was the Wiktionary project which has IPA phonetic transliterations available and is much more comprehensive in terms of the number of transliteration records and languages available. A version of this has been curated in the Wikipron project that has been used in various SIGMORPHON workshops. However, for the purposes of an initial investigation the smaller ipa-dict dataset was chosen as we found it exemplary of many languages in an easy to access way.

Given the ipa-dict data, we pre-processed the dataset by removing punctuation, splitting tokens by individual characters, and separated each individual dataset by language. Removing punctuation and splitting tokens by individual characters is problematic from a phonetic perspective since punctuation can change pronunciation and individual phonemes can be composed of multiple characters for instance in the case of gnocchi, an Italian pasta, whose phonetic transliteration /ˈɲɔk.ki/ transforms the "gn" onto the voiced palatal nasal /ɲ/ and ch into the voiceless velar plosive /k/. Similarly, the comma in English can transform how adjacent words are blended during speech. In the case of tokenizing IPA transcriptions, we split units that have a single phonetic meaning such as the case of diacritics and other components that modify the preceding letter. However, we don't consider these major obstacles for two reasons. In the tokenization case, we are looking to model transliteration from a sequence-to-sequence perspective where multiple characters can be decoded as single phonemes or single characters can be decoded as multiple phonemes depending on the context of proceeding characters and language under transliteration. For punctuation, since we are focusing here on word-to-word transliterations we don't consider punctuation a major feature in this task.

Our aim to model transliteration is to look at how the process of transliteration can be learned across languages. Therefore, we split the dataset into the 22 languages which have over 10k transliteration records. We select a random sample of 10k transliteration records for each language. By focusing on a 10k set, we are emulating the medium resource sample size given in previous SIGMORPHON conferences [1]. We do this so that each transliteration model has the same number of samples to train from. In order to train the model that we describe below, we split the dataset into a training set of 8k samples, an evaluation test set of 1k samples, and a final test set of 1k samples.

The architecture we chose to model transliteration was the popular transformer architecture [7]. We chose the transformer architecture because it provides a proven approach to model sequence transduction where positional context is considered across the sequence using attention and self-attention that has been proven out in both character level [8] and transliteration contexts [9] achieving state-of-the-art g2p performance. The transliteration process can be understood as sequence transduction where a source language orthography for a given word is translated into a target language orthography. With attention, we theorize that we will be able to adequately translate a phonetic letter given both previous and preceding characters in the input sequence. Additionally, with self-attention, we theorize that the decoder can learn additional phonetic transliteration procedures that ensure that certain phonetic sequences are more likely than others. For instance, learning to attend to certain sequence dependencies like vowel harmony. Finally, with multi-head attention and multiple layers we theorize that multiple levels of phonetic and orthographic features can be captured in understanding the input sequence and producing the output sequence.

Table 1 describes the parameters used to train a transformer model using the architecture presented in [7]. The learning rate was set to a uniform rate of 0.005 using an Adam optimizer and dropout rate of 0.1 was used. The implementation of Vaswani [7] was made in PyTorch and the code is available on request. One thing to note is that the transformers used are relatively shallow and trained for a relatively low duration of 20 epochs. The choice of a shallow model was due to the relatively limited amount of training data and speed at which the model was able to converge on a relatively stable loss for each language. Larger models were underfitting and slow to converge. However, in the discussion below we theorize that this could potentially be resolved with self-supervision pretraining or a larger dataset such as Wiktionary. A large batch of 512 was chosen due to [8] which suggests that for character-level transformers batch size is critical especially in the case of low resource settings where transliteration tasks often take place.

Table 1. Parameters used to train each transliteration model.

| Dropout Rate | Learning Rate | Batch Size | Embedding dimension | Layers | Feed Forward Size | Attention Size | Number of Attention Heads |
|---|---|---|---|---|---|---|---|
| 0.1 | 0.005 | 512 | 32 | 2 | 32 | 32 | 2 |

## 4. Results

We trained 22 transliteration models in order to compare the ability to learn transliteration for different languages. Once the models were trained, they were evaluated on their respective test sets using a simple character-level accuracy metric as seen in [10]. The accuracy metric for each predicted sequence is the mean of per-token comparison of the predicted IPA transliteration and the ground truth transliteration. Simple character level transliteration was chosen over the more common word and phoneme error rate (See [9]) for demonstrative purposes in this paper rather than to prove performance against some baseline. In future studies, word and phoneme error rate could be more illustrative of performance and provide standard comparative metrics. In Table 2, we present these accuracies across each language we trained a transliteration model for. Notably, Esperanto, Malay, Swahili, and Vietnamese are the most accurately transliterated while Cantonese, Japanese, and Mandarin (hans indicates simplified and hant indicates traditional) are the least accurately transliterated. In order to explore these results further we had the following questions on what might contribute to the differences of accuracies.

Following [4], we theorized that the range of letters used by the source language, it's graphemes, and the range of IPA letters used, its phonemes, are potentially illustrative of the complexity involved in the language. We also show the ratio of grapheme to phonemes used including a convenience column that shows the distance from a 1:1 ratio. These numbers are presented below in Table 2 with German and French (Quebec) appearing to use the most IPA sounds and Malay and Esperanto using the least. The smallest source alphabets were Swahili, English (both UK and US), and Malay and the largest were Mandarin. As for the ratio of sounds used to source alphabet length, German, English (UK), and Swahili had the highest ratio of sounds used per letter with Esperanto, French (France), Spanish (Both Mexico and Spain) having the closest to a 1:1 ratio and Mandarin and Cantonese having the lowest ratio. Figure 1 explores these results graphically presenting the ratio of IPA letters used to source language letters used plotted against the accuracy of each model. One clear limitation of this approach is that a simple ratio of grapheme to phoneme does not robustly indicate how close a language is to representing phonemes in a 1:1 way. An example of this is Arabic, which appears in relatively close alignment but in this dataset is not using any graphemes for vowels as is common in standard written Arabic.

Table 2. Transliteration model for languages, their vocabulary length, and ratio of number of IPA tokens used to source vocabulary tokens used.

| Language | Accuracy | IPA Vocab Length | Source Vocab Length | Ratio of IPA to Source Length | Distance from 1:1 ratio |
|---|---|---|---|---|---|
| Mandarin (hant) | **71.98%** | 41 | **23283** | 0.002 | **1.00** |
| Mandarin (hans) | 73.88% | 41 | 20505 | 0.002 | 1.00 |
| Cantonese | 79.95% | 33 | 14672 | 0.002 | 1.00 |
| Japanese | 73.66% | 32 | 5510 | 0.006 | 0.99 |
| Vietnamese (Southern) | 95.53% | 43 | 90 | 0.478 | 0.52 |
| Vietnamese (Northern) | 95.65% | 43 | 90 | 0.478 | 0.52 |
| Vietnamese (Central) | 96.32% | 45 | 90 | 0.500 | 0.50 |
| Odia | 95.00% | 38 | 63 | 0.603 | 0.40 |
| Arabic | 87.10% | 32 | 38 | 0.842 | 0.16 |
| Esperanto | 97.08% | 27 | 29 | 0.931 | 0.07 |
| French (France) | 92.12% | 43 | 46 | 0.935 | 0.07 |
| Spanish (Mexico) | 95.68% | 32 | 33 | 0.970 | 0.03 |
| Spanish (Spain) | 94.85% | 33 | 33 | 1.000 | 0.00 |
| Malay | 96.87% | 30 | 27 | 1.111 | 0.11 |
| Finnish | 91.81% | 38 | 34 | 1.118 | 0.12 |
| French (Quebec) | 90.80% | 54 | 47 | 1.149 | 0.15 |
| Norwegian | 84.74% | 48 | 34 | 1.412 | 0.41 |
| English (US) | 80.30% | 37 | 26 | 1.423 | 0.42 |
| Swedish | 85.42% | 48 | 33 | 1.455 | 0.45 |
| Swahili | 96.63% | 40 | 24 | 1.667 | 0.67 |
| English (UK) | 83.92% | 47 | 26 | 1.808 | 0.81 |
| German | 85.55% | **84** | 32 | **2.625** | **1.63** |

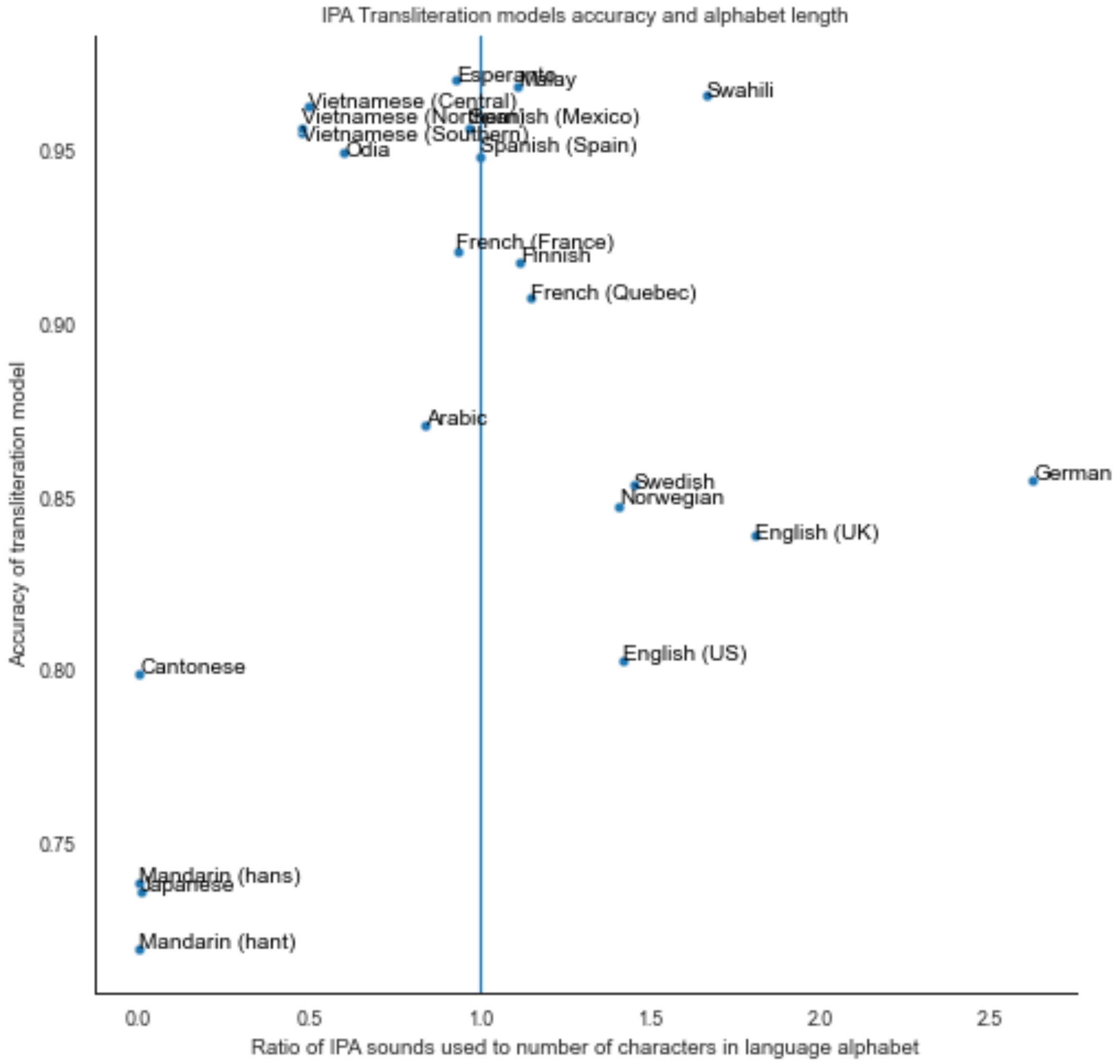

Figure 1. IPA Transliteration model accuracy and the ratio of IPA length to source language alphabet. The blue line indicates a 1:1 ratio for grapheme-to-phoneme.

## 5. Discussion

When looking at learning to transliterate we are interested in two core competencies. The first is learning to read, deciphering a source alphabet and making some transformation of it, and the second is learning to pronounce, understanding how a source alphabet maps to phonemes. In the above experiments, we are asking how learning to read and pronounce under the g2p task differ across languages. Why does Mandarin appear the hardest to learn and Esperanto, a constructed language, the easiest? The above results provide an empirical justification for two intuitive hypotheses presented below: logographic writing systems are harder to learn to transliterate than phonetic writing systems, second that languages with high redundancy for phonemes or close to a 1:1 ratio of letter to phoneme are the easiest to learn.

First, we should say that the ratio of grapheme-to-phoneme is a very simplistic measure of language complexity, as is demonstrated in the case of Arabic noted above. However, we feel justified in using it at the very least to illustrate a few things about language complexity because it allows us to look at the size of an input vocabulary (the graphemes) and output vocabulary (the phonemes) as the model being trained would. That is without explicit knowledge about how to evaluate the complexity of a language. We will see this below when we discuss the notion of complexity as it relates to source vocabulary size and data sparsity. Secondly, the below discussions are based on. Another related limitation that should be presented before discussing results is that Latin-script based orthographies are overrepresented in our comparison. Table 3 shows that we include 4 logographic orthographies, 1 abugida, 1 abjad, and 16 Latin-based orthographies. In future works, we should try to study more languages that do not use Latin-based orthographies.

Given the limitations above, the major discrepancies in accuracy results are the differences between languages with logographic writing and languages without. This is an intuitive result and follows naturally from the fact that the number of values the input alphabet can take is orders of magnitudes larger than non-logographic orthographies. If each unique character is understood as a class under a multi-class classification scheme, then we can easily see that some languages are much sparser than others. Summarizing results from the SIGMORPHON 2020 workshop [1] discusses the lack of performance in Korean. The authors note that some orthographies, notably Korean in their dataset, are much sparser than others. Meaning the mapping between the number of training samples and source graphemes is very low or sparse compared to other languages. In order to illustrate the dynamics of this in our dataset we have tabulated the sparsity of each language we are using in Table 3 which shows the number of samples per unique character per language. In this case Mandarin only has 0.34 examples per letter and Japanese only has 1.45 samples per class. The next lowest number of samples per letter in a source language alphabet is Vietnamese with 88 samples per class. We know that the accuracy of a classifier will naturally depend on the number of training samples per class producing underfitting in a situation without enough data. However, we can't simply say that large source vocabularies require more training data to learn transliteration as Vietnamese is one of the easiest languages to learn, achieving an accuracy of 96.32% on Vietnamese (Central) despite only having 88 samples per letter. (Interestingly, Vietnamese was also the lowest baseline word error rate in the SIGMORPHON 2020 task [1]). Again, English (US) has the lowest accuracy score of 80.30% among non-logographic orthography despite having 307 samples per letter. Based on this we can observe that logographic orthographies are harder to learn to transliterate than phonetic ones. We should be wary to suggest that there is a linear relationship between number of letters in a source vocabulary and its ease of learning, at least in the case where non-logographic orthographies are involved.

Table 3. Samples per unique character in alphabet for each language given a training set of 8000 tokens.

| Language | Type of Orthography | Accuracy | Number of Unique Characters | Number of Samples per Unique Character |
|---|---|---|---|---|
| Mandarin (hant) | Logographic | 71.98% | 23283 | 0.34 |
| Mandarin (hans) | Logographic | 73.88% | 20505 | 0.39 |
| Cantonese | Logographic | 79.95% | 14672 | 0.54 |
| Japanese | Logographic | 73.66% | 5510 | 1.45 |
| Vietnamese (Southern) | Latin based | 95.53% | 90 | 88.88 |
| Vietnamese (Northern) | Latin based | 95.65% | 90 | 88.88 |
| Vietnamese (Central) | Latin based | 96.32% | 90 | 88.88 |
| Odia | Abugida | 95.00% | 63 | 126.98 |
| French (Quebec) | Latin based | 90.80% | 47 | 170.21 |
| French (France) | Latin based | 92.12% | 46 | 173.91 |
| Arabic | Abjad | 87.10% | 38 | 210.52 |
| Finnish | Latin based | 91.81% | 34 | 235.29 |
| Norwegian | Latin based | 84.74% | 34 | 235.29 |
| Spanish (Mexico) | Latin based | 95.68% | 33 | 242.42 |
| Spanish (Spain) | Latin based | 94.85% | 33 | 242.42 |
| Swedish | Latin based | 85.42% | 33 | 242.42 |
| German | Latin based | 85.55% | 32 | 250 |
| Esperanto | Latin based | 97.08% | 29 | 275.86 |
| Malay | Latin based | 96.87% | 27 | 296.29 |
| English (US) | Latin based | 80.30% | 26 | 307.69 |
| English (UK) | Latin based | 83.92% | 26 | 307.69 |
| Swahili | Latin based | 96.63% | 24 | 333.33 |

In order to explain why certain languages are easier to learn than others we hypothesize that first and intuitively the phonetic alphabets are easiest to learn. Among those languages with close to a 1:1 mapping between a grapheme and a single phoneme or with many redundancies in terms of multiple graphemes mapping to a single phoneme, or which are highly expressive with respect to how source letters can be transliterated are the easiest to learn. In Figure 1, we see those languages with close to a 1:1 ratio of grapheme to phoneme appear to do the best. Languages which are more expressive in terms of a phonetic writing system, meaning they have multiple and even redundant graphemes to phonemes, also appear to do well. Languages that appear to do poorly like English and German have a low ratio of letters in the source language to phonemes they express. This means that the languages have letters that can map to multiple different phonemes introducing which would have the effect of introducing ambiguity in the g2p task.

Based on the findings above, we can propose an initial hypothesis that there appears to be a pattern where languages with the closest grapheme-to-phoneme ratio or language with the most expressive phonetic orthographies tend to be the easiest to learn. Related to this, we should note that under the transliteration task cross-lingual comparison of languages appears to be constrained by data sparsity. The ease of learning a language from a computational point of view is tied to the number of samples we can observe for each transliteration pair. Languages with large source orthographies, especially logographic ones, are dramatically more sparse than other ones. With this in mind, we should question the fairness of comparing languages under g2p transliteration tasks where data sparsity is dramatically different. Perhaps this can explain some of the results on why logographic languages appear to be harder to learn. We speculate that if each language was given a training set proportional to their source vocabulary, then we could have a much more robust way to compare the complexity of languages under the machine learning scheme. We urge future authors who are using joint orthography and phonology datasets such as those studying g2p to consider the sparsity of data when comparing languages. Additionally, future work should try to design tasks where the relative data sparsity is fairer.

## 6. Conclusion

In this work we have suggested that exploring the process of transliteration can yield interesting insights into the process of language learning especially for machine models. In order to further pursue this work further a few interesting notes should be made. First, as seen with logographic languages and languages with high phoneme-source letter ambiguity the number of training samples must be higher for effective learning therefore in future work more comprehensive datasets should be used such as Wiktionary and comparison based not on absolute number of samples but on number of samples per class should be made. Additionally, in order to prove out the theory that the simplest (closest to a 1:1 mapping between phoneme and grapheme) or most expressive phonetic orthographies are the easiest to pronounce we should sample many different orthographies and devise a metric for complexity that isn't simply a ratio of grapheme over IPA letter. Due to the simplicity of the ratio of absolute number in grapheme and phoneme inventories, future work should develop measures based on theorical results from linguistics on how each phoneme maps to each grapheme. Finally, once a fairer complexity measure and task is developed for cross-lingual comparison, we should look at performing an ablation study that illustrates how complexity of learning to pronounce a language change under training and modeling strategies that are known to improve the ability to transliterate such as encoding phonetic similarity in features, self-supervision, data augmentation, and joint or cross language modeling.

**Authors**

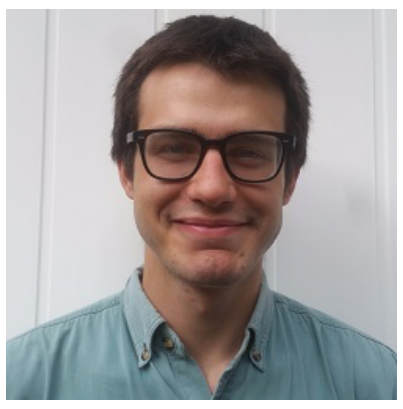

Domenic Rosati, MIS (Dalhousie 2017) is a research scientist at scite.ai. His interests include investigating foundational aspects of natural language understanding and reasoning.